\documentclass[letterpaper, 10 pt, conference]{ieeeconf}  

\IEEEoverridecommandlockouts                              

\usepackage{graphics} 
\usepackage{epsfig} 
\usepackage{mathptmx} 
\usepackage{times} 
\usepackage{amsmath} 
\usepackage{amssymb}  
\usepackage{subfig}
\usepackage{cite}
\usepackage[amssymb]{SIunits}
\usepackage{graphicx}
\usepackage{amssymb}
\usepackage{multirow}
\usepackage{wrapfig}
\usepackage{seqsplit}

\usepackage{amsmath}
\usepackage{bm} 
\usepackage{footnote}
\usepackage{threeparttable}
\usepackage{epsfig}
\usepackage{graphicx}
\usepackage{amsmath}
\usepackage{balance}

\usepackage[utf8]{inputenc} 
\usepackage[T1]{fontenc}    
\usepackage{url}            
\usepackage{booktabs}       
\usepackage{amsfonts}       
\usepackage{nicefrac}       
\usepackage{microtype}      
\usepackage{lipsum}
\usepackage{paralist}
\usepackage{color}
\usepackage{aecompl}

\makeatletter
\let\NAT@parse\undefined
\makeatother
\usepackage[colorlinks = true,
            linkcolor = blue,
            urlcolor  = blue,
            citecolor = blue,
            anchorcolor = blue]{hyperref}

\usepackage{array}
\newcommand{\PreserveBackslash}[1]{\let\temp=\\#1\let\\=\temp}
\newcolumntype{C}[1]{>{\PreserveBackslash\centering}p{#1}}
\newcolumntype{R}[1]{>{\PreserveBackslash\raggedleft}p{#1}}
\newcolumntype{L}[1]{>{\PreserveBackslash\raggedright}p{#1}}

\newcommand{\rom}[1]{\uppercase\expandafter{\romannumeral #1\relax}}

\newcommand{\fref}[1]{Fig.~\ref{#1}}
\newcommand{\sref}[1]{Section~\ref{#1}}
\newcommand{\tref}[1]{Table~\ref{#1}}

\newcommand{\etal}{\textit{et al.}~}
\newcommand{\ie}{{i.e.},~}

\title{\LARGE \bf
AirVO: An Illumination-Robust Point-Line Visual Odometry
}

\author{Kuan Xu$^{1}$, Yuefan Hao$^{2}$, Shenghai Yuan$^{1}$, Chen Wang$^{3}$, Lihua Xie$^{1}$, \emph{Fellow, IEEE} 
\thanks{*This research is supported by the National Research Foundation, Singapore under its Medium Sized Center for Advanced Robotics Technology Innovation.}
\thanks{$^{1}$Kuan Xu, Shenghai Yuan, and Lihua Xie are with School of Electrical and Electronic Engineering, Nanyang Technological University, 50 Nanyang Avenue, Singapore 639798,
        {\tt\small \{kuan.xu,shyuan,elhxie\}@ntu.edu.sg}}%
\thanks{$^{2}$Yuefan Hao is with the Robot R\&D Department, Geekplus Corp., Beijing 100107, China
        {\tt\small yuefan.hao@outlook.com}}%
\thanks{$^{3}$Chen Wang is with the Spatial AI \& Robotics Lab at The Department of Computer Science \& Engineering, State University of New York at Buffalo, Buffalo, NY 14260, USA
        {\tt\small chenw@sairlab.org}}%
}

\begin{document}

\maketitle
\thispagestyle{empty}
\pagestyle{empty}

\begin{abstract}



This paper proposes an illumination-robust visual odometry (VO) system that incorporates both accelerated learning-based corner point algorithms and an extended line feature algorithm.
To be robust to dynamic illumination, the proposed system employs the convolutional neural network (CNN) and graph neural network (GNN) to detect and match reliable and informative corner points. 
Then point feature matching results and the distribution of point and line features are utilized to match and triangulate lines. 
By accelerating CNN and GNN parts and optimizing the pipeline, the proposed system is able to run in real-time on low-power embedded platforms.
The proposed VO was evaluated on several datasets with varying illumination conditions, and the results show that it outperforms other state-of-the-art VO systems in terms of accuracy and robustness. The open-source nature of the proposed system allows for easy implementation and customization by the research community, enabling further development and improvement of VO for various applications.

\end{abstract}

\section{INTRODUCTION}

Due to the good balance in cost and accuracy,
VO has been used in an extensive range of applications, especially in the domain of augmented reality and robotics \cite{macario2022comprehensive}. 
Despite the existence of numerous well-known works, such as  MSCKF \cite{mourikis2007multi}, VINS-Mono \cite{qin2018vins} and OKVIS \cite{leutenegger2015keyframe}, 
the existing solutions are not robust enough for illumination-challenging conditions \cite{cadena2016past}. For example, in dynamic illumination environments, visual tracking becomes more challenging and thus the quality of the estimated trajectory is severely affected \cite{zuniga2020vi}. 

On the other hand, deep learning technology has made great progress in many computer vision tasks, which has triggered another research trend \cite{xu2022aircode}. A lot of learning-based feature extraction and matching methods have been proposed and they have been proven to be more robust than handcrafted methods in illumination-challenging environments \cite{yi2016lift, detone2018superpoint, sarlin2020superglue}.
However, they often require huge computational resources and thus are impractical for real-time applications with lightweight robotics platforms such as unmanned aerial vehicles.


\begin{figure}[th]
    \centering
    \hspace{-0.5cm} \includegraphics[width=0.95\linewidth]{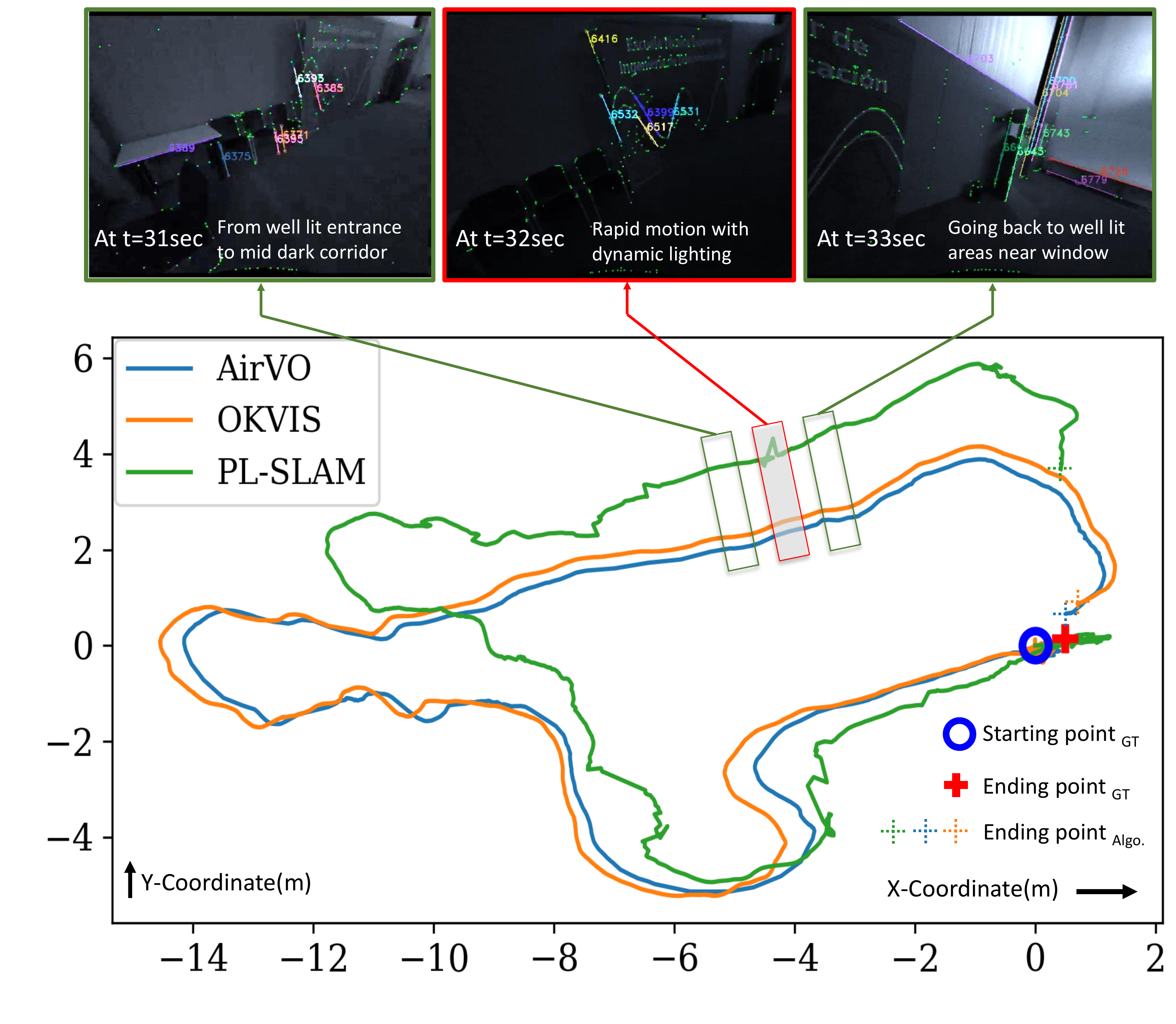}
    \caption{AirVO is an accurate and robust stereo visual odometry in illumination-challenging environments. More demos are available at \url{https://youtu.be/YfOCLll_PfU}.}
    \label{fig:uma-traj}
    \vspace{-1.5em}
\end{figure}

Therefore, in this paper, we propose AirVO, an illumination-robust stereo visual odometry. We employ both learning-based feature extraction and matching methods to make our system robust enough in illumination-challenging environments. To achieve real-time and cost-effective performance, we accelerate the CNN and GNN parts and optimize the pipeline, making the feature extraction and tracking five times faster than the original work and the whole system able to run at a rate of about 15Hz on a low-power embedded device.

 To improve the accuracy, we also introduce line features into our system. We argue that long lines can provide more stable and accurate constraints, so we merge the short lines detected by LSD \cite{von2012lsd}. However, line detection is usually unstable in dynamic illumination environments, which makes line tracking and matching more difficult than in good lighting conditions. Besides, line feature triangulation is more difficult than point feature triangulation, because it suffers more from degenerate motion \cite{yang2019visual}.
Therefore, we also propose a fast and efficient illumination-robust line processing pipeline in this paper. Observing that the point tracking in our system is very robust, we associate points with lines according to their distances. Then lines can be matched and triangulated using the matching and triangulation results of related points. The proposed line processing method is shown to be very robust even when the line detection is not stable and the lighting conditions are challenging. It is also very fast due to that it does not need to extract line descriptors. Overall, our contributions are as follows:
\begin{itemize}

        \item The key contribution in this paper is that we propose a novel hybrid VO system that can effectively handle varying illumination conditions. Our proposed system combines the efficiency of traditional optimization techniques with the robustness of learning-based methods. To our best knowledge, AirVO is the first visual odometry that employs both learning-based feature detection and matching algorithms and can run in real-time on low-power embedded platforms.
        \item We propose a new line processing pipeline for VO in this paper. Our approach associates 2D lines with learning-based 2D points on the image, leading to more robust feature matching and triangulation. This novel method enhances the accuracy and reliability of VO, especially in illumination-challenging environments.
        \item We perform extensive experiments that prove the efficiency and effectiveness of the proposed methods. The results show that AirVO outperforms other state-of-the-art VO and visual-inertial odometry (VIO) systems, especially in illumination-challenging environments. Through optimization and acceleration of the relevant parts, our point feature detection and matching achieve more than 5$\times$ faster than the original work. Additionally, the system can run at a rate of about 15Hz on a low-power embedded device and 40Hz on a notebook PC. We release source code at \url{https://github.com/sair-lab/AirVO} to benefit the community.

\end{itemize}

\begin{figure*}[t]
    \vspace{0.5em}
    \centering
    \includegraphics[width=1\linewidth]{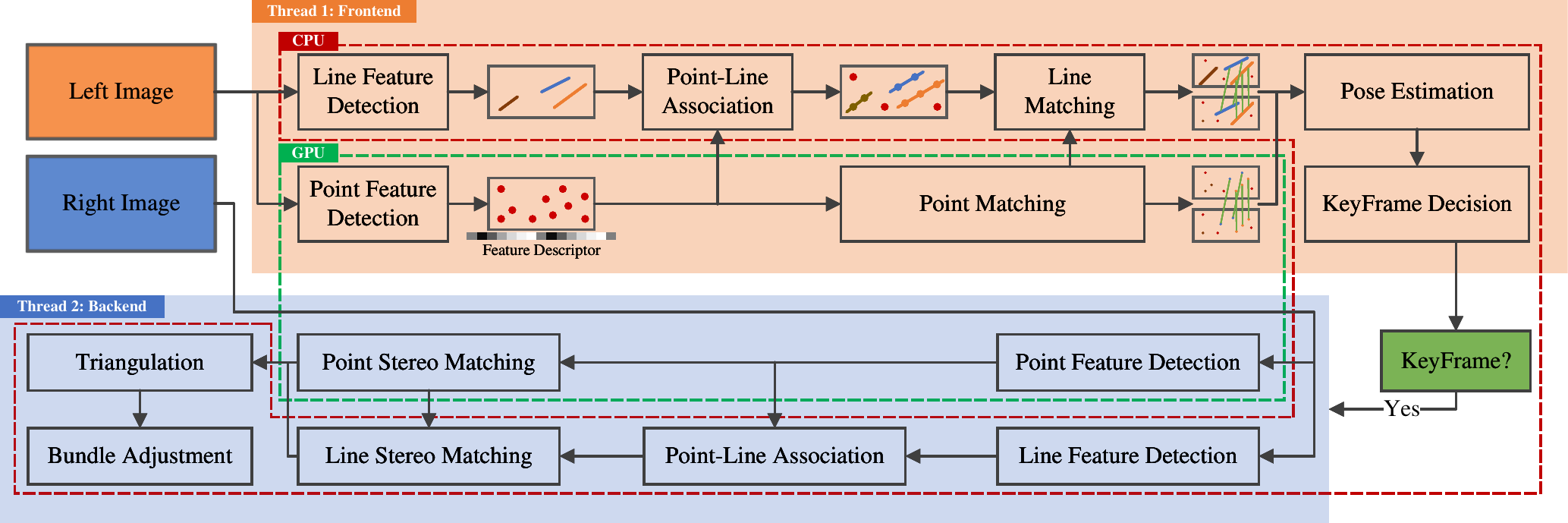}
    \caption{The framework of AirVO. The system is split into two main threads, which are represented by two different colored regions. The modules in the red dotted box and green dotted box run on the CPU and GPU, respectively. 
    }
    \label{fig:pipeline}
    \vspace{-1.9em}
\end{figure*}

\section{Related Works}

\subsection{Feature Extraction and Tracking for Visual SLAM}

Various key-point features have been proposed and applied to different computer vision tasks. Many of these features, e.g., ORB \cite{rublee2011orb}, FAST \cite{viswanathan2009features}, and BRISK \cite{leutenegger2011brisk} are applied to VO and SLAM systems, e.g., ORB-SLAM \cite{mur2015orb}, VINS-Mono\cite{qin2018vins}, because of their balanced effectiveness and efficiency. Two methods are widely used to track the feature points. The first is to use optical flow \cite{qin2018vins}, and the other is matching by descriptor \cite{mur2017orb, mourikis2007multi}. However, most of the current visual SLAM systems based on the above methods are evaluated in well-lighted environments and make a brightness consistency assumption. Thereby, their performances are significantly affected by challenging lighting conditions, such as dark, over-bright or dynamic illumination conditions.

With the development of deep learning techniques, many learning-based feature extraction and matching methods have been proposed and started to be applied to visual SLAM. Kang \etal \cite{kang2019df} introduce TFeat network \cite{balntas2016learning} to extract descriptors for FAST corners in a traditional VSLAM pipeline. Tang \etal \cite{tang2019gcnv2} use a neural network to extract robust key-points and binary feature descriptors with the same shape of the ORB. Han \etal \cite{han2020superpointvo}  combine SuperPoint \cite{detone2018superpoint} feature extractor with a traditional back-end. Bruno \etal proposed LIFT-SLAM \cite{bruno2021lift}, where they use LIFT \cite{yi2016lift} to extract features. Li \etal \cite{li2020dxslam} replace ORB feature with SuperPoint in ORB-SLAM2 and optimize the feature extraction with Intel OpenVINO toolkit. However, the above methods still adopt traditional methods to track or match these learning-based features, making them not robust enough to changing illumination. Sarlin \etal propose HF-Net \cite{sarlin2019coarse}, where they integrate SuperPoint and SuperGlue \cite{sarlin2020superglue} into COLMAP \cite{schonberger2016structure}, a structure from motion software. HF-Net achieves good performance for visual place recognition tasks but requires huge computing resources and is unable to build maps in real-time. 

Unlike current methods, AirVO introduces both learning-based feature extraction and matching methods in the VO system, which makes our system robust enough in illumination-challenging environments. By accelerating the CNN and GNN parts, our system can perform pose estimation and build maps in real-time on low-power platforms.

\subsection{Line Matching for Visual SLAM}

Line features widely exist in man-made environments, which can provide additional constraints. One of the challenges of using lines in visual SLAM is to perform line matching. A method used in many current SLAM systems \cite{gomez2019pl, zuo2017robust, yang2019visual, lim2022uv} is to match lines through LBD \cite{zhang2013efficient} descriptor. This method may make the line matching fail as the traditional line detection method, such as LSD \cite{von2012lsd}, may be unstable. To overcome this, some systems \cite{zhou2021dplvo, zou2019structvio, xu2022leveraging} sample some points on a line, and then track the line by tracking these points. However, using either minimizing photometric error along the epipolar line or zero-normalized cross-correlation (ZNCC) matching method \cite{di2005zncc} cannot ensure robust line tracking in dynamic illumination environments. 

\subsection{Visual SLAM for Dynamic Illumination}

Several methods have been proposed to improve the robustness of VO and Visual SLAM to illumination changes. 
DSO \cite{engel2017direct} models brightness changes and jointly optimizes camera poses and photometric parameters. 
DRMS \cite{gu2021drms} and AFE-ORB-SLAM \cite{yu2022afe} utilize various image enhancements. 
Some systems try different methods, such as ZNCC, locally-scaled sum of squared differences (LSSD) and dense descriptor computation, to achieve robust tracking \cite{scandaroli2012improving, crivellaro2014robust, usenko2019visual}. 
These methods mainly focus on either global or local illumination change for all kinds of images, however, lighting conditions often affect the scene differently in different areas \cite{park2017illumination}.

Other related methods include that of Huang and Liu \cite{huang2019robust}, which presents a multi-feature extraction algorithm to extract two kinds of image features when a single-feature algorithm fails to extract enough feature points. Kim \etal \cite{kim2019autonomous} employ a patch-based affine illumination model during direct motion estimation. Chen \etal \cite{chen2021robust} minimize the normalized information distance (NID) with nonlinear least square optimization for image registration. Alismail \etal propose a binary feature descriptor using a descriptor assumption to avoid the brightness constancy \cite{alismail2016direct}.

\section{Methodology}

\subsection{System Overview} 

The proposed framework is shown in \fref{fig:pipeline}. It is a hybrid VO system where we utilize both the learning-based front end and the traditional optimization backend. For each stereo image pair, we first employ SuperPoint \cite{detone2018superpoint} to extract feature points on the left image and match them with the last keyframe using SuperGlue \cite{sarlin2020superglue}, and in parallel, we also extract line features. Then the two kinds of features are associated according to their distances and line features are matched using the matching results of associated points. 
After that, we perform an initial pose estimation and reject outliers. Based on the results, we select keyframes, extract features on the right image and triangulate 2D points and lines of keyframes. Finally, the local bundle adjustment will be performed to optimize points, lines and keyframe poses.

To improve system efficiency, We replace the 32-bit floating-point arithmetic of CNNs and GNNs in our system with 16-bit floating-point arithmetic, which makes feature extraction and tracking more than five times faster than the original code on the embedded device. We also design a multi-thread pipeline that utilizes both CPU and GPU resources. A producer-consumer model is used to split the system into two main threads, i.e., the feature thread and the optimization thread. In the feature thread, we use two sub-threads to process point features and line features separately. In one sub-thread, the point feature extraction and matching with the last frame are put on the GPU while in parallel, the other sub-thread is used to extract line features on the CPU. In the optimization thread, we perform initial pose estimation and keyframe decision. If a new keyframe is selected, we extract both point and line features on its right image and optimize its pose with a local map. 
 

\subsection{2D Line Processing}

We first give the details of 2D line processing in our system, which includes line segment detection and matching. 

\subsubsection{Detection}\label{sec:2d_line_detection}

Line detection of AirVO is based on a traditional method (\ie LSD \cite{von2012lsd}) for efficiency. LSD is a popular line detection algorithm. However, it suffers from the problem of dividing a line into multiple segments. Therefore, we improve it by merging two line segments $\mathbf{l}_1$ and $\mathbf{l}_2$ if the following conditions are satisfied:
\begin{itemize}
    \item The angle between $\mathbf{l}_1$ and $\mathbf{l}_2$ is less than a threshold  $\delta_\theta$.
    \item The distance between the midpoint of one line and the other line is not greater than a certain value $\delta_\mathbf{d}$.
    \item If projections of $\mathbf{l}_1$ and $\mathbf{l}_2$ on $\mathbf{X}$-coordinate axis and $\mathbf{Y}$-coordinate axis do not have overlap, the distance of the two closest endpoints is smaller than a threshold $\delta_\mathbf{ep}$.
\end{itemize}

\begin{figure}[t]
    \vspace{0.5em}
    \centering
    \includegraphics[width=0.95\linewidth]{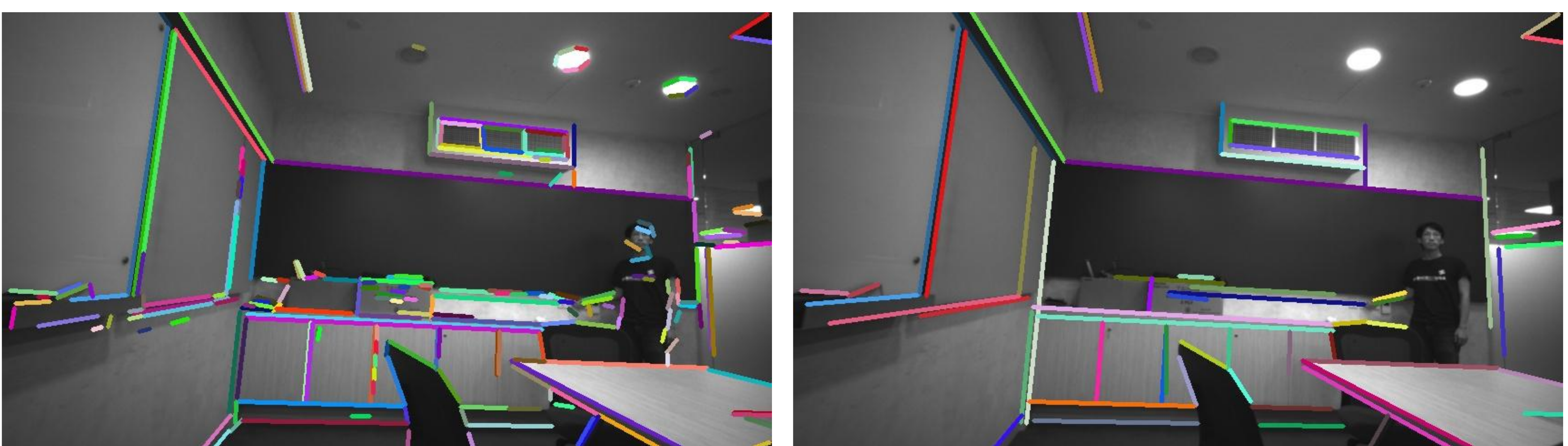}
    \caption{Lines detected by LSD (left) and by AirVO (right). We merge unstable short lines into stable longer lines.}
    \label{fig:line_detection}
    \vspace{-1.8em}
\end{figure}

The line features detected in our system and comparison with LSD are shown in \fref{fig:line_detection}.
We argue that long line segments are more repetitive and less affected by noise than the short ones, so after the merger, line segments whose lengths are less than a preset threshold will be filtered out so that only long line segments are used in the following stages.


\subsubsection{Matching}\label{sec:2d_line_matching}

Most of the current VO and SLAM systems use LBD algorithm or tracking sample points to match or track lines. 
LBD algorithm extracts the descriptor from a local band region of the line, so it suffers from unstable line detection in dynamic illumination environments where the line length may change and thus the local band region would be different between two frames. Tracking sample points can track the line which has different lengths in two frames, but current SLAM systems usually use optical flow to track the sample points, which have a bad performance when the light conditions change rapidly or violently. Some learning-based line feature matching methods \cite{pautrat2021sold2, pautrat2022deeplsd} are also proposed, however, they are rarely used in current SLAM systems as a result of the requirement for huge computational resources. We do not employ them either because it is difficult to make the system run in real-time on low-power embedded platforms if both learning-based point features and learning-based line features are used simultaneously.

Therefore, to address both the effectiveness problem and efficiency problem, we design a fast and robust line-matching method for dynamic illumination environments. First, we associate point features with line segments through the distances between points and lines. Assume that $M$ key-points and $N$ line segments are detected on the image, where each point is denoted as $\mathbf{p}_i=(x_i,y_i)$ and each line segment is denoted as $\mathbf{l}_j=(A_j,B_j,C_j,x_{j,1},y_{j,1}, x_{j,2}, y_{j,2})$, where $(A_j,B_j,C_j)$ are line parameters of $\mathbf{l}_j$ and $(x_{j,1},y_{j,1}, x_{j,2}, y_{j,2})$ are the endpoints. We first compute the distance between $\mathbf{p}_i$ and $\mathbf{l}_j$ through:
\begin{equation}\label{eq:point_line_distance}
   d_{ij} = d \left(\mathbf{p}_i, \mathbf{l}_j \right) = \frac{\lvert A_j \cdot x_i + B_j \cdot y_i + C_j \rvert}{\sqrt{A_j^2 + B_j^2}} .
\end{equation}
If $d_{ij} < 3$ and the projection of $\mathbf{p}_i$ on the coordinate axis lies within the projections of line segment endpoints, i.e., $ \min({x}_{j,1}, {x}_{j,2}) \leq x_i \leq \max({x}_{j,1}, {x}_{j,2})$ or $ \min({y}_{j,1}, {y}_{j,2}) \leq {y}_i \leq \max({y}_{j,1}, {y}_{j,2})$, we will say $\mathbf{p}_i$ belongs to $\mathbf{l}_j$. Then the line segments on two images can be matched based on the point-matching result of these two images. For ${^k\mathbf{l}_m}$ on image $k$ and ${^{k+1}\mathbf{l}_n}$ on image $k+1$, we compute a score to represent the confidence of that they are the same line:
\begin{equation}\label{eq:line_matching_score}
   {S}_{mn} = \frac{{N}_{pm}}{\min({^k{N}_m}, {^{k+1}{N}_n})},
\end{equation}
where ${N}_{pm}$ is the matching number between point features belonging to ${^k\mathbf{l}_m}$ and point features belonging to ${^{k+1}\mathbf{l}_n}$. ${^k{N}_m}$ and ${^{k+1}{N}_n}$ are the numbers of point features belonging to ${^k\mathbf{l}_m}$ and ${^{k+1}\mathbf{l}_n}$, respectively. Then if ${S}_{mn} > \delta_{S}$ and ${N}_{pm} > \delta_{N}$, where $\delta_{S}$ and $\delta_{N}$ are two preset thresholds, we will regard ${^k\mathbf{l}_m}$ and ${^{k+1}\mathbf{l}_n}$ as the same line.

\begin{figure}[t]
    \vspace{0.5em}
    \centering
    \includegraphics[width=0.99\linewidth]{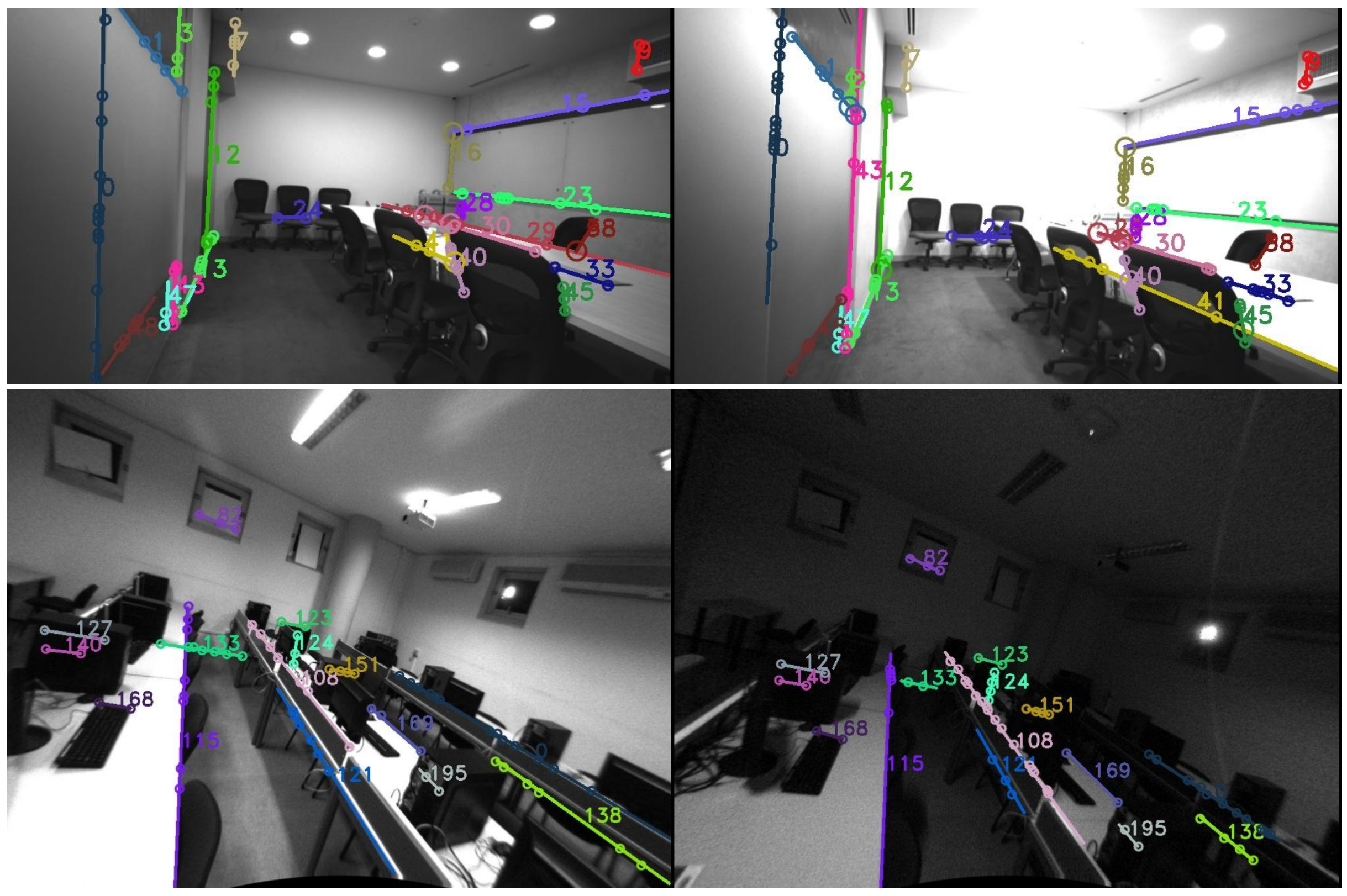}
    \caption{Line matching of AirVO in challenging scenes. Matched lines are drawn in the same color. Circles on a line are the points associated with the line. A larger radius indicates that the point is associated with more lines.}
    \label{fig:line_matching}
    \vspace{-1.5em}
\end{figure}

Because the point matching is illumination-robust and feature association is not affected by lighting changes, the proposed line tracking method is very robust to dynamic illumination environments, as shown in \fref{fig:line_matching}. 

\subsection{3D Line Processing}

In this part, we will introduce our 3D line processing methods. Compared with 3D points, 3D lines have more degrees of freedom, so we first introduce their representations in different stages. Then the methods of 
line triangulation, \ie constructing a 3D line from some 2D line segments, and line re-projection, \ie projecting the 3D line to the image plane, will be illustrated in detail.

\subsubsection{Representation}\label{sec:3d_line_representation}
We use Pl\"{u}cker coordinates \cite{bartoli2005structure} to represent a 3D spatial line:
\begin{equation}\label{eq:line_plucker}
    \mathbf{L} = \left[\begin{array}{c} \mathbf{n} \\ \mathbf{v} \end{array}\right] \in \mathbb{R}^{6},
\end{equation}
where $\mathbf{v}$ is the direction vector of the line and $\mathbf{n}$ is the normal vector of the plane determined by the line and the origin. Pl\"{u}cker coordinates are used for 3D line triangulation, transformation, and projection to the image. It is over-parameterized because it is a 6-dimensional vector, but a 3D line has only four degrees of freedom. In the graph optimization stage, the extra degrees of freedom will increase the computational cost and cause the numerical instability of the system \cite{zuo2017robust}. Therefore, we also use orthonormal representation \cite{bartoli2005structure} to represent a 3D line:
\begin{equation}\label{eq:line_orthonormal}
    \left( \mathbf{U}, \mathbf{W} \right) \in SO(3) \times SO(2)
\end{equation}

The relationship between Pl\"{u}cker coordinates and orthonormal representation is similar to $SO(3)$ and $so(3)$. Orthonormal representation can be obtained from Pl\"{u}cker coordinates by:
\begin{equation}\label{eq:line_plucker_orthonormal_conversion}
    \mathbf{L} = \left[ \mathbf{n} \mid \mathbf{v} \right] = \underbrace{ \left[\begin{array}{ccc} \frac{\mathbf{n}}{\lVert \mathbf{n} \lVert} & \frac{\mathbf{v}}{\lVert \mathbf{v} \lVert} & \frac{\mathbf{n} \times \mathbf{v}}{\lVert \mathbf{n} \times \mathbf{v} \lVert}  \end{array}\right]}_{\mathbf{U} \in SO(3)} \underbrace{\left[\begin{array}{cc} \lVert \mathbf{n} \lVert & \\ & \lVert \mathbf{v} \lVert \\ & \end{array}\right]}_{\Sigma_{3 \times 2}},
\end{equation}
where $\Sigma_{3 \times 2}$ is a diagonal matrix and its two non-zero entries defined up to scale can be represented by an $SO(2)$ matrix, i.e., $\mathbf{W}$. In practice, this conversion can be done simply and quickly with the QR decomposition.

\subsubsection{Triangulation}\label{sec:3d_line_triangulation}
Triangulation is to initialize a 3D line from two or more 2D line observations. Two methods are used to triangulate a 3D line in our system. The first is similar to the line triangulation algorithm $B$ in \cite{yang2019visual}, where a 3D line can be computed from two planes. To achieve this, we select two line segments, $\mathbf{l}_1$ and $\mathbf{l}_2$, on two images, which are two observations of a 3D line. $\mathbf{l}_1$ and $\mathbf{l}_2$ can be back-projected and construct two 3D planes, $\mathbf{\pi}_1$ and $\mathbf{\pi}_2$. Then the 3D line can be regarded as the intersection of $\mathbf{\pi}_1$ and $\mathbf{\pi}_2$.

However, triangulating a 3D line is more difficult than triangulating a 3D point, because it suffers more from degenerate motions\cite{yang2019visual}. Therefore, we also employ a second line triangulation method if the above method fails, where points are utilized to compute the 3D line. In \sref{sec:2d_line_matching}, we have associated point features with line features. So to initialize a 3D line, two triangulated points $\mathbf{X}_1$ and $\mathbf{X}_2$, which belong to this line and have the shortest distance from this line on the image plane are selected. Then the Pl\"{u}cker coordinates of this line can be obtained through:
\begin{equation}\label{eq:line_two_points_plucker_conversion}
    \mathbf{L} = \left[\begin{array}{c} \mathbf{n} \\ \mathbf{v} \end{array}\right] = \left[\begin{array}{c} \mathbf{X}_1 \times \mathbf{X}_2 \\ \frac{\mathbf{X}_1 - \mathbf{X}_2}{\lVert \mathbf{X}_1 - \mathbf{X}_2 \lVert} \end{array}\right].
\end{equation}
Because the selected 3D points have been triangulated in the point triangulating stage, this method requires little extra computation. It is very efficient and robust.

\begin{table*}[t]
    \vspace{0.5em}
    \setlength\tabcolsep{3.5pt}
    \caption{Translational error (RMSE) without loop closing and re-localization on the OIVIO dataset (unit: m). \textcolor{red}{L} refers to tracking lost and \textcolor{red}{D} refers to sequences where RMSEs are larger than 10m. }
    \label{tab:oivio_rmse}
    \centering  
    \begin{tabular}{C{0.11\linewidth}C{0.1\linewidth}C{0.1\linewidth}C{0.09\linewidth}C{0.08\linewidth}C{0.08\linewidth}C{0.1\linewidth}C{0.1\linewidth}C{0.08\linewidth}}
        \toprule
         Sequence & VINS-Fusion & StructVIO & UV-SLAM & PL-SLAM & OKVIS & ORB-SLAM2 & Basalt-VIO  & AirVO  \\ \midrule
        MN\_015\_GV\_01 &     0.1033         &    8.6098          &   0.4991           & 1.3166 & \underline{0.0663} & 0.0762 & 0.2157              & \textbf{0.0537}  \\
        MN\_015\_GV\_02 & \textcolor{red}{D} & \textcolor{red}{L} & \textcolor{red}{D} & 0.9523 & 1.5320             & \underline{0.0776} & 0.1533  & \textbf{0.0619} \\
        MN\_050\_GV\_01 & \textcolor{red}{D} & \textcolor{red}{D} & \textcolor{red}{D} & 1.1538 & 0.7785             & \underline{0.0839} & 0.1857  & \textbf{0.0756} \\
        MN\_050\_GV\_02 & \textcolor{red}{D} & \textcolor{red}{D} & \textcolor{red}{D} & 1.0055 & 0.7385             & \underline{0.0755} & 0.1026  & \textbf{0.0717} \\
        MN\_100\_GV\_01 & \textcolor{red}{D} & \textcolor{red}{D} & \textcolor{red}{D} & 0.8455 & 0.8729             & \underline{0.0892} & 0.1965  & \textbf{0.0646} \\
        MN\_100\_GV\_02 & \textcolor{red}{D} & \textcolor{red}{D} & \textcolor{red}{D} & 0.6708 & 0.4360             & \underline{0.0848} & 0.0922  & \textbf{0.0770}  \\
        TN\_015\_GV\_01 &     0.1541         &    7.5849          &   1.6695           & 1.8856 & 0.3063             & \textbf{0.0902}    & 0.1478  & \underline{0.1009}  \\
        TN\_050\_GV\_01 &     0.2079         & \textcolor{red}{D} &   2.5948           & 1.9335 & 0.2262             & \textbf{0.0965}    & 0.5214  & \underline{0.0971}  \\
        TN\_100\_GV\_01 &     0.4063         & \textcolor{red}{D} &   1.4496           & 1.5263 & 0.3984             & \underline{0.1044} & 0.1162  & \textbf{0.0578} \\
        \bottomrule
    \end{tabular}
    \vspace{-1.7em}
\end{table*}    


\subsubsection{Re-projection}\label{sec:3d_line_projection}

We use Pl\"{u}cker coordinates to transform and re-project 3D lines. First, we convert the 3D line from the world frame to the camera frame:
\begin{equation}\label{eq:line_transformation}
    ^c\mathbf{L} = \left[\begin{array}{c} ^c\mathbf{n} \\ ^c\mathbf{v} \end{array}\right] = \left[\begin{array}{cc} {_w^c\mathbf{R}} & \left[{_w^c\mathbf{t}} \right]_{\times}{_w^c\mathbf{R}} \\ \mathbf{0} & {_w^c\mathbf{R}} \end{array}\right]\left[\begin{array}{c} ^w\mathbf{n} \\ ^w\mathbf{v} \end{array}\right] = {_w^c\mathbf{H}} {^w\mathbf{L}},
\end{equation}
where $^c\mathbf{L}$ and $^w\mathbf{L}$ are Pl\"{u}cker coordinates of 3D line in the camera frame and world frame, respectively. ${_w^c\mathbf{R}} \in SO(3)$ is the rotation matrix from world frame to camera frame and ${_w^c\mathbf{t}} \in \mathbb{R}^{3}$ is the translation vector. $\left[\cdot \right]_{\times}$ denotes the skew-symmetric matrix of a vector and ${_w^c\mathbf{H}}$ is the transformation matrix of 3D lines from world frame to camera frame. 

Then the 3D line $^c\mathbf{L}$ can be projected to the image plane through a line projection matrix ${_c^i\mathbf{P}}$:
\begin{equation}\label{eq:line_re-projection}
    ^i\mathbf{l} = \left[\begin{array}{c} A \\ B \\ C \end{array}\right] = {_c^i\mathbf{P}}{^c\mathbf{L}}_{[:3]}  = \left[\begin{array}{ccc} f_x & 0 & 0 \\ 0 & f_y & 0 \\ -f_yc_x & -f_xc_y & f_xf_y \end{array}\right] {^c\mathbf{n}},
\end{equation}
where $^i\mathbf{l} = \left[\begin{array}{ccc} A & B & C \end{array}\right]^T$ is the re-projected 2D line on image plane. ${^c\mathbf{L}}_{[:3]}$ donates the first three rows of vector $^c\mathbf{L}$. 

\subsection{Keyframe Selection}

Observing that the learning-based data association method used in our system is able to track two frames that have a large baseline, so different from the frame-by-frame tracking strategy used in other VO or visual SLAM systems, we only match the current frame with the last keyframe, as this can reduce the tracking error. A frame will be selected as a keyframe if any of the following conditions is satisfied:

\begin{itemize}
    \item The distance to the last keyframe is larger than $\delta_d^{kf}$.
    \item The angle with the last keyframe is larger than $\delta_\theta^{kf}$.
    \item The number of tracked map-points is smaller than $N_1^{kf}$ and bigger than $N_2^{kf}$, where $N_2^{kf} < N_1^{kf}$.
    \item Tracked map-points are more than $N_2^{kf}$ but the tracking-lost happened in the last frame, i.e., map-points tracked by the last frame are less than $N_2^{kf}$.
\end{itemize}
where $\delta_d^{kf}$, $\delta_\theta^{kf}$, $N_1^{kf}$ and $N_2^{kf}$ are all preset thresholds.

\subsection{Graph Optimization}

We select $N_{kf}^{go}$ keyframes and construct a co-visibility graph similar to ORB-SLAM \cite{mur2015orb}, where map points, 3D lines, and keyframes are vertices and constraints are edges. Both point constraints and line constraints are used in our system and the related error terms are defined as follows. 

\subsubsection{Line Re-projection Error}\label{sec:mono_line_constraint}
If the frame $k$ can observe the 3D line $^w\mathbf{L}_i$, then the re-projection error is defined as:
\begin{subequations}\label{eq:mono_line_constraint}
    \begin{align}\label{eq:line_error}
        \mathbf{E}_{l_{k, i}} &= e_l \left({^k\bar{\mathbf{l}}_i}, {_c^k\mathbf{P}} \left({{_w^c\mathbf{H}} {^w\mathbf{L}_i}} \right)_{[:3]} \right) \in \mathbb{R}^{2} ,\\
        e_l \left({^k\bar{\mathbf{l}}_i}, ^k\mathbf{l}_i \right) &= \left[\begin{array}{cc} d \left(^k\bar{\mathbf{p}}_{i, 1}, ^k\mathbf{l}_i \right) & d \left(^k\bar{\mathbf{p}}_{i, 2}, ^k\mathbf{l}_i \right) \end{array}\right]^T,
    \end{align}
\end{subequations}
where ${^k\bar{\mathbf{l}}_i}$ is the observation of $^w\mathbf{L}_i$ on frame $k$, $d \left(\mathbf{p}, \mathbf{l} \right)$ is the distance between point $\mathbf{p}$ and line $\mathbf{l}$, and ${^k\bar{\mathbf{p}}_{i, 1}}$ and ${^k\bar{\mathbf{p}}_{i, 2}}$ are the endpoints of ${^k\bar{\mathbf{l}}_i}$. 


\subsubsection{Point Re-projection Error}\label{sec:mono_point_constraint}
If the frame $k$ can observe the 3D point $^w\mathbf{X}_q$, then the re-projection error is defined as:
\begin{equation}\label{eq:mono_point_constraint}
    \mathbf{E}_{p_{k, q}} = {^k\bar{\mathbf{x}}_q} - \pi \left( {_w^c\mathbf{R}} {^w\mathbf{X}_q} + {_w^c\mathbf{t}} \right),
\end{equation}
where ${^k\bar{\mathbf{x}}_q}$ is the observation of $^w\mathbf{X}_q$ on frame $k$ and $\pi \left(\cdot \right)$ represents the camera projection.

\begin{figure}[t]
    \vspace{0.5em}
    \centering
    \includegraphics[width=0.82\linewidth]{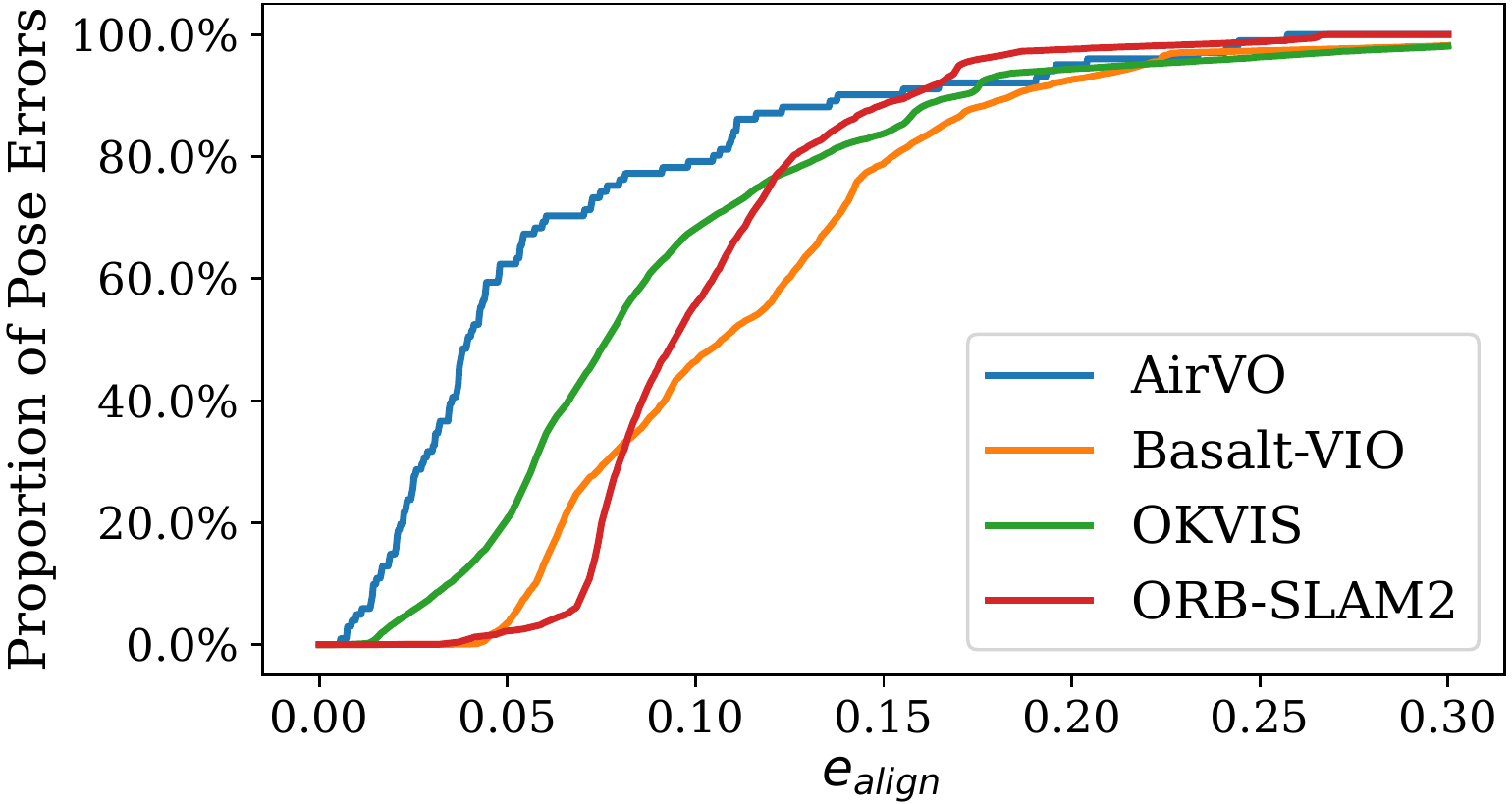}
    \caption{Comparison based on the OIVIO dataset. The vertical axis is the proportion of pose errors that are less than the given alignment error threshold on the horizontal axis.}
    \label{fig:oivio-rmse-curve}
    \vspace{-1.8em}
\end{figure}

\begin{figure*}[t]
    \vspace{0.5em}
    \centering
    \includegraphics[width=0.98\linewidth]{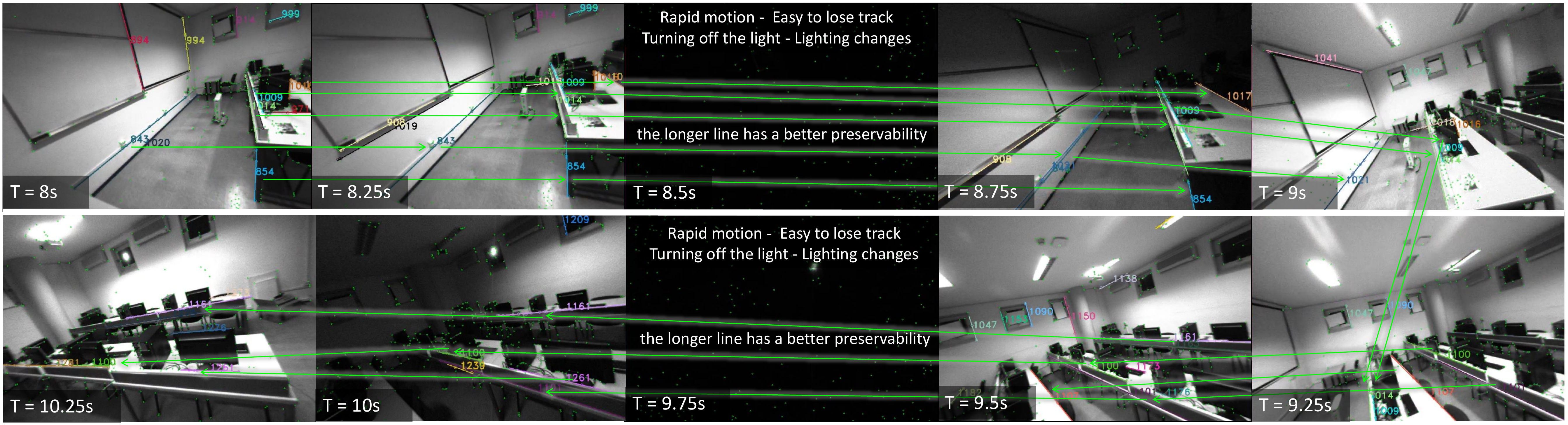}
    \caption{A challenging sequence in UMA-VI dataset with significant illumination changes. The image may suddenly go dark as a result of turning off the lights, which is very difficult for feature tracking.}
    \label{fig:uma-seq}
    \vspace{-1.8em}
\end{figure*}

\section{EXPERIMENTS}

In this section, experimental results will be presented to demonstrate the performance of our method. We take pre-trained SuperPoint and SuperGlue to detect and match feature points without any fine-tuning training. The experiments are conducted on two datasets: OIVIO dataset \cite{kasper2019benchmark} and UMA visual-inertial dataset \cite{zuniga2020vi}. To prove the efficiency of the proposed line processing pipeline, we compare AirVO with state-of-the-art point-line VO and visual SLAM systems, i.e., PL-SLAM \cite{gomez2019pl}, StructVIO\cite{zou2019structvio} and UV-SLAM \cite{lim2022uv}. PL-SLAM and UV-SLAM use the LBD descriptor to match line features while StructVIO tracks line features by tracking sampling points on lines. We also add stereo-mode VINS-Fusion \cite{qin2018vins}, ORB-SLAM2 \cite{mur2017orb}, Basalt-VIO \cite{usenko2019visual} and VIO-mode OKVIS \cite{leutenegger2013keyframe} to the baselines. To handle the dynamic illumination problem, VINS-Fusion uses a failure detection and recovery module, while StructVIO uses the ZNCC method and Basalt-VIO uses the LSSD KLT method. 
The following experiments will prove that AirVO outperforms these methods in illumination-challenging environments. As the proposed method is a VO system, we disabled the loop closure part and re-localization from the above baselines.

Note that as a result of lacking the ground truth of line matching and triangulation in dynamic illumination environments, 
we prove the effectiveness of the proposed line processing method by comparing it with other point-line systems and an ablation study instead of designing an extra line matching or triangulation comparison. 
Like \cite{cvivsic2022soft2,song2023contextavo, zhang2022towards}, we compare with ORB-SLAM2 instead of ORB-SLAM3 \cite{campos2021orb} because the newly added atlas and IMU in ORB-SLAM3 are unfair to compare with the visual-only odometry, and they are difficult to remove because of the high coupling system.
We also do not add DX-SLAM \cite{li2020dxslam} and GCNv2-SLAM \cite{tang2019gcnv2} to the baselines since they are based on RGB-D inputs and thus cannot run on the stereo datasets.

\begin{table}[t]
    \caption{Translational error (RMSE) on the UMA-VI dataset (unit: m). The best results are \textbf{highlighted}. }
    \label{tab:uma_rmse}
    \centering
    \begin{tabular}{C{0.25\linewidth}C{0.14\linewidth}C{0.12\linewidth}C{0.12\linewidth}}
        \toprule
         Sequence          & PL-SLAM & OKVIS  & AirVO            \\ \midrule
        conference-csc1    &  2.6974 & 1.1181 & \textbf{0.5236}  \\ 
        conference-csc2    &  1.5956 & 0.4696 & \textbf{0.1607}  \\
        third-floor-csc1   &  4.4779 & 0.2525 & \textbf{0.1760}  \\
        third-floor-csc2   &  6.0675 & 0.2161 & \textbf{0.1312}  \\
        average            &  3.7096 & 0.5141 & \textbf{0.2479}  \\
        \bottomrule
    \end{tabular}
    \vspace{-1.7em}
\end{table}

\subsection{Results on OIVIO Benchmark} \label{sec:oivio_test}

OIVIO dataset collects visual-inertial data in tunnels and mines. In each sequence, the scene is illuminated by an onboard light of approximately 1300, 4500, or 9000 lumens. We used all nine sequences with ground truth acquired by the Leica TCRP1203 R300. The performance of translational error is presented in \tref{tab:oivio_rmse}. The two most accurate results are \textbf{highlighted} and \underline{underlined}, respectively. AirVO achieves the best performance on 7 sequences and the second-best performance on the other 2 sequences, which outperforms other state-of-the-art algorithms. We notice that VINS-Fusion, StructVIO and UV-SLAM lost track on many sequences, this may be because their feature tracking methods, \ie LSSD KLT sparse optical flow, ZNCC, LBD descriptor, are not robust enough in illumination-challenging environments. 


\begin{table}[t]
    \caption{Ablation study. Translational error (RMSE) of AirVO$^{\text{w/o line}}$ and AirVO on the UMA-VI and OIVIO datasets (unit: m). The best results are \textbf{highlighted}.}
    \label{tab:ablation}
    \centering
    \begin{tabular}{C{0.12\linewidth}C{0.25\linewidth}C{0.16\linewidth}C{0.12\linewidth}}
        \toprule
        & Sequence           & AirVO$^{\text{w/o line}}$ & AirVO \\ \midrule
        \multirow{4}{*}{UMA-VI} & 
          conference-csc1    & 2.4789             & \textbf{0.5236}  \\ 
        & conference-csc2    & 0.2323             & \textbf{0.1607}   \\
        & third-floor-csc1   & \textbf{0.1736}    & 0.1760  \\
        & third-floor-csc2   & 0.1629             & \textbf{0.1312}  \\
        \specialrule{0em}{1pt}{1pt}
        \hline   
        \specialrule{0em}{1pt}{1pt}
        \multirow{9}{*}{OIVIO} &
          MN\_015\_GV\_01 &  0.1035              & \textbf{0.0537} \\
        & MN\_015\_GV\_02 &  0.0668              & \textbf{0.0619} \\
        & MN\_050\_GV\_01 &  0.1051              & \textbf{0.0756} \\
        & MN\_050\_GV\_02 &  0.1049              & \textbf{0.0717}          \\
        & MN\_100\_GV\_01 &  0.1177              & \textbf{0.0646} \\
        & MN\_100\_GV\_02 &  0.0921              & \textbf{0.0770} \\
        & TN\_015\_GV\_01 &  0.1155              & \textbf{0.1009} \\
        & TN\_050\_GV\_01 &  0.0987              & \textbf{0.0971} \\
        & TN\_100\_GV\_01 &  0.0879              & \textbf{0.0578}         \\
        \bottomrule
    \end{tabular}
    \vspace{-1.7em}
\end{table}

We show a comparison of our method with selected baselines on OIVIO TN\_100\_GV\_01 sequence in \fref{fig:oivio-rmse-curve}. In this case, the robot goes through a mine with onboard illumination. The distance is about 150 meters and the average speed is about 0.84m/s. The plot shows the proportion of pose errors on the horizontal axis that are less than the given alignment error threshold on the horizontal axis. AirVO achieves the most accurate result on this sequence.

\subsection{Results on UMA-VI Benchmark}

UMA-VI dataset is a visual-inertial dataset gathered in illumination-challenging scenarios with handheld custom sensors. We selected sequences with illumination changes to evaluate our system. As shown in \fref{fig:uma-seq}, it contains many sub-sequences where the image suddenly goes dark as a result of turning off the lights. It is more challenging than OIVIO dataset, so we only select methods proved to be illumination-robust in \sref{sec:oivio_test} as baselines, \ie ORB-SLAM2, PL-SLAM, OKVIS and Basalt-VIO. The translational errors are presented in \tref{tab:uma_rmse}. As ORB-SLAM2 and Basalt-VIO lost track on all 4 sequences, we do not list their results. It can be seen that AirVO outperforms other methods. Its average translational error is only 6.7\% of PL-SLAM and 48.2\% of OKVIS. We notice that the aligned errors are larger than those on the OIVIO dataset. It is because the UMA-VI dataset only gives the ground truth of the beginning and end of each sequence, which makes the errors appear larger, and the scenes are more difficult for VO or VIO systems.



We also compare the trajectory of AirVO with OKVIS and PL-SLAM on conference-csc2 sequence as shown in \fref{fig:uma-traj}. The traveling distance of this sequence is about 50 meters and the average speed is about 0.75m/s. It clearly shows that AirVO produces the best accuracy in this challenging case. The drift error of AirVO is about 1.0\%. OKVIS and PL-SLAM are 1.5\% and 7.1\%, respectively. 

\subsection{Ablation Study}

To show the effectiveness of the proposed line processing method, we remove line features from AirVO and name it as AirVO$^{\text{w/o line}}$. The comparison results of AirVO and AirVO$^{\text{w/o line}}$ on OIVIO and UMA-VI datasets are presented in \tref{tab:ablation}. It can be seen that AirVO outperforms AirVO$^{\text{w/o line}}$ on 12 of 13 sequences, and utilizing line features reduces the translational error by 58.2\% on average, which demonstrates that the proposed line processing method can improve the performance of the system. 

\subsection{Runtime Analysis}

This section presents the running time analysis of the proposed system. The evaluation is performed on the Nvidia Jetson AGX Xavier (2018), which is a low-power embedded platform with an 8-core ARM v8.2 64-bit CPU and a low-power 512-core NVIDIA Volta GPU. The resolution of the input image sequence is 640 × 480. For all algorithms, we extract 200 points and disabled the loop closure, re-localization and visualization part for a fair comparison.

\subsubsection{CNN and GNN Acceleration}
We first verify the acceleration of the point detection and matching network. In our system, detecting and tracking feature points for one image take 64 \milli\second, while the original code needs 342 \milli\second. So it is about 5.3$\times$ faster than the origin.

\begin{figure}[t]
    \vspace{0.5em}
    \centering
    \includegraphics[width=0.9\linewidth]{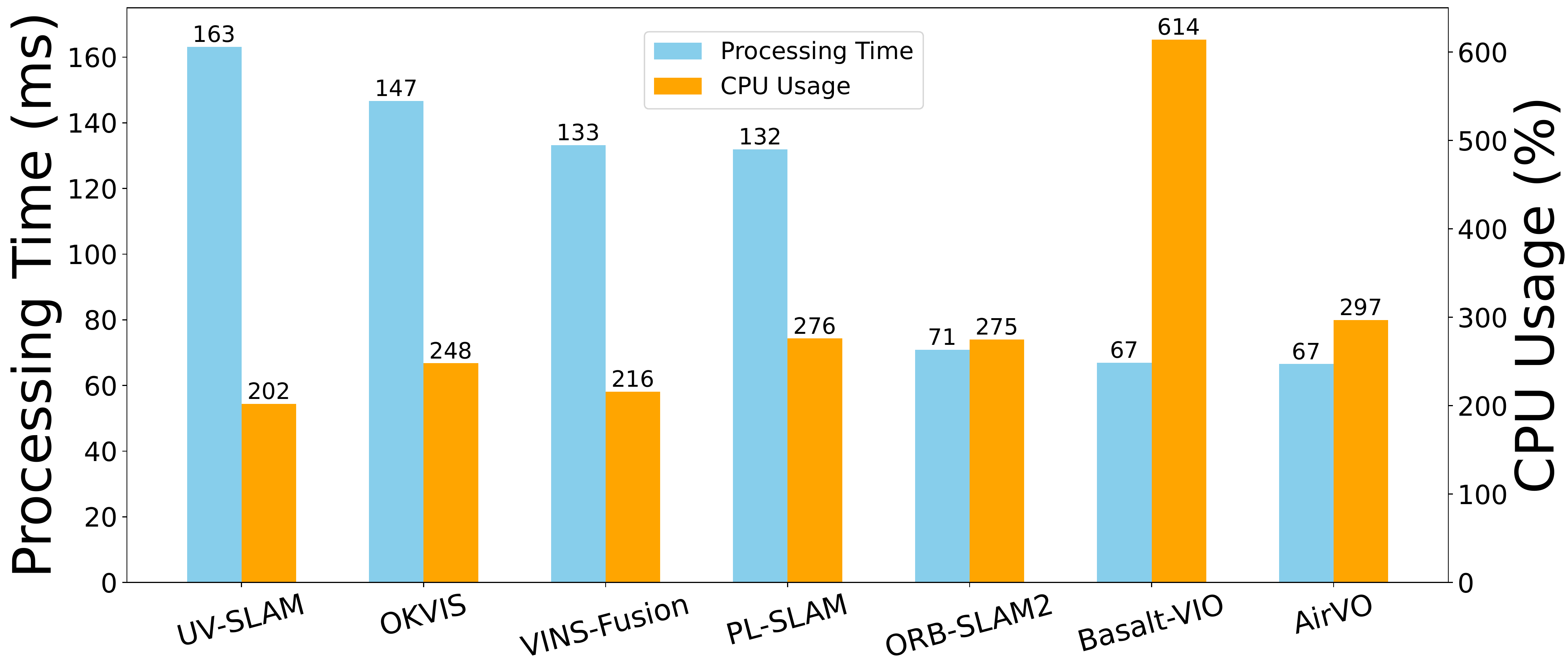}
    \caption{Bar chat showing the efficiency of different algorithms, as measured by CPU usage (\%) and per-frame processing time (ms) on Nvidia Jetson AGX Xavier (2018).}
    \label{fig:time}
    \vspace{-0.7em}
\end{figure}

\begin{table}[t]
    \caption{The average running time comparison of principal components with PL-SLAM. }
    \label{tab:time_detail}
    \centering
    \begin{tabular}{C{0.14\linewidth}C{0.08\linewidth}C{0.08\linewidth}C{0.08\linewidth}C{0.08\linewidth}C{0.08\linewidth}C{0.1\linewidth}}
        \toprule
                   & PE   & LE   & PM   & LM   & IPE   & BA   \\ \midrule
        PL-SLAM    & 70 \milli\second & 96 \milli\second & 1 \milli\second  & 29 \milli\second & 1 \milli\second   & 101 \milli\second  \\ 
        AirVO      & 25 \milli\second & 29 \milli\second & 10 \milli\second & 2 \milli\second  & 4 \milli\second   & 264 \milli\second  \\
        \bottomrule
    \end{tabular}
    \vspace{-1.5em}
\end{table}

\subsubsection{Efficiency Comparison}
We also compare the algorithm efficiency, as measured by CPU usage and per-frame processing time.  The result is presented in \fref{fig:time}. It can be seen that AirVO is one of the fastest methods (about 15 FPS) while the CPU usage is roughly the same as other methods because of utilizing the GPU resources. Notice that only the binary executable file of Struct-VIO is available, which is compiled on the x86 computer and cannot run on the Jetson platform, so we did not add it to this comparison.

\subsubsection{Detailed Running Time}
We also present the detailed running time of each module of PL-SLAM and AirVO in \tref{tab:time_detail}, where PE is point extraction, LE is line extraction, PM is point matching, LM is line matching, IPE is initial pose estimation and BA is keyframe processing and local bundle adjustment. It can be seen that the line processing pipeline of AirVO is much more efficient than PL-SLAM.
Our BA module has higher runtime than PL-SLAM, this may be because more stable features are detected, tracked and taken into optimization in our system.
Notice that these modules run in parallel and BA module is a non-real-time back-end thread, so the running time of the whole system is not the simple accumulation of each module.

\section{CONCLUSIONS}

In this work, we presented an illumination-robust visual odometry based on learning-based key-point detection and matching methods. To improve the accuracy, line features are also utilized in our system. We proposed a novel line processing pipeline to make line tracking robust enough in illumination-dynamic environments. In the experiments, we showed that the proposed method achieved superior performance in illumination-dynamic environments and could run in real-time on low-power devices. We open the source code to benefit the robotic community. For future work, we will extend AirVO to a SLAM system by adding loop closing, re-localization and map reuse. We hope to build an illumination-robust visual map for long-term localization.








{
    \balance
    \bibliographystyle{IEEEtran}
    \bibliography{papers}
}

\end{document}